\documentclass[letterpaper]{article} 
\usepackage{aaai24}  
\usepackage{times}  
\usepackage{helvet}  
\usepackage{courier}  
\usepackage[hyphens]{url}  
\usepackage{graphicx} 
\usepackage{natbib}  
\usepackage{caption} 
\usepackage{algorithm}
\usepackage{algorithmic}

\usepackage{newfloat}
\usepackage{listings}
\DeclareCaptionStyle{ruled}{labelfont=normalfont,labelsep=colon,strut=off} 
\floatstyle{ruled}
\newfloat{listing}{tb}{lst}{}
\floatname{listing}{Listing}
\usepackage{hyperref}

\usepackage{amsmath,booktabs,multirow,array,xcolor,makecell,subcaption,tikz,pgfplotstable,graphbox,pifont,pgf-pie,amsfonts,microtype}
\pgfplotsset{compat=1.18}

\makeatletter
\newcommand*{\@rowstyle}{}
\newcommand*{\rowstyle}[1]{
  \gdef\@rowstyle{#1}%
  \@rowstyle\ignorespaces%
}
\newcolumntype{=}{
  >{\gdef\@rowstyle{}}%
}
\newcolumntype{+}{
  >{\@rowstyle}%
}
\makeatother

\title{MeDM: Mediating Image Diffusion Models for Video-to-Video Translation with Temporal Correspondence Guidance}
\author{
    Ernie Chu, Tzuhsuan Huang, Shuo-Yen Lin, Jun-Cheng Chen
}
\affiliations{
    Research Center for Information Technology Innovation, Academia Sinica \\

    128 Academia Road, Section 2 \\
    Nankang, Taipei 115, Taiwan (R.O.C.) \\
    \{shchu, jason890425, joseph, pullpull\}@citi.sinica.edu.tw
}

\begin{document}

\maketitle

\begin{abstract}
This study introduces an efficient and effective method, MeDM, that utilizes pre-trained image Diffusion Models for video-to-video translation with consistent temporal flow. The proposed framework can render videos from scene position information, such as a normal G-buffer, or perform text-guided editing on videos captured in real-world scenarios. We employ explicit optical flows to construct a practical coding that enforces physical constraints on generated frames and mediates independent frame-wise scores. By leveraging this coding, maintaining temporal consistency in the generated videos can be framed as an optimization problem with a closed-form solution. To ensure compatibility with Stable Diffusion, we also suggest a workaround for modifying observation-space scores in latent Diffusion Models. Notably, MeDM does not require fine-tuning or test-time optimization of the Diffusion Models. Through extensive qualitative, quantitative, and subjective experiments on various benchmarks, the study demonstrates the effectiveness and superiority of the proposed approach. Our project page can be found at \url{https://medm2023.github.io/}.
\end{abstract}

\section{Introduction}
Recently, diffusion-based deep generative models (DMs) \cite{ho2020ddpm, song2021ddim, song2021score} have set new benchmarks in image generation, particularly for text-to-image tasks. Major tech companies are investing heavily in large-scale text-to-image models, like DALL·E 2 \cite{ramesh2022dalle2}, Imagen \cite{saharia2022imagen}, and Stable Diffusion \cite{rombach2022LDM}. These models deliver high-quality images within seconds based on text descriptions. Additionally, models like InstructPix2Pix \cite{brooks2023instructpix2pix} and ControlNet \cite{zhang2023controlnet} enable instruction-based and structure-preserving text-guided image-to-image translations. These advances have spurred novel applications in content creation. However, generating videos or performing video-to-video translation while maintaining temporal consistency remains a challenging and ongoing problem. 

To tackle this challenge, many text-to-video models have been proposed, including Imagen Video~\cite{ho2022imagenvideo}, Make-A-Video~\cite{singer2022make-a-video}, etc. However, these approaches demand substantial training videos and computational resources. To mitigate these requirements, researchers have extended text-to-image models to encompass video generation, incorporating a temporal aspect into the latent Diffusion Models (LDMs)~\cite{blattmann2023videoldm}. In addition, several zero-shot methods are also proposed to directly leverage the pre-trained image-based or text-to-image-based DMs for text-guided video-to-video translation, including Pix2Video~\cite{ceylan2023pix2video}, Rerender A Video~\cite{yang2023rerender}, ControlVideo~\cite{chen2023control-a-video}, etc. While these methods achieve video transformation, some still experience persistent flickering issues, while others involve intricate and meticulous processes.

\begin{figure*}[t]
    \centering
    \includegraphics[width=\linewidth]{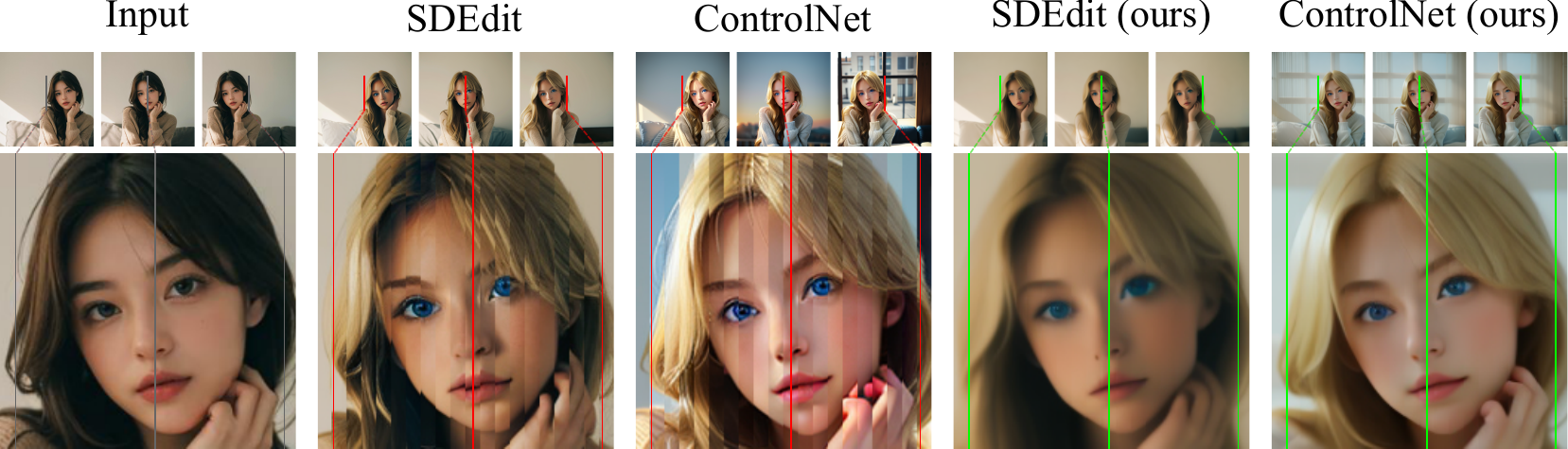}
    \caption{\textbf{Text-guided video editing.} From a series of video frames (top row), we extract a 20-pixel-wide vertical segment of pixels from each generated frame and stack them horizontally (bottom). A fluent video should reconstruct a stripe-free image. From left to right: input video (synthetic motion), results from SDEdit \cite{meng2022sdedit} and ControlNet \cite{zhang2023controlnet}, before and after applying the proposed harmonization framework. For SDEdit, we start the denoising process from $0.6T$ step. For ControlNet, we start from the pure noise and use the lineart \cite{chan2022lineart} derived from the input as conditions. Prompt: \textit{1girl, soft light, blonde hair, blue eyes.}}
    \label{fig:teaser}
\end{figure*}

To perform efficient and high-quality temporally consistent video-to-video translation, we propose a novel temporal correspondence guidance based on optical flows to Mediate image Diffusion Models (MeDM, pronounced “medium”) for video-to-video translation. Explicit optical flows are utilized to establish an encoding based on pixel correspondences across video frames. By leveraging this encoding, pixels on a trajectory stay consistent to their corresponding pixel in a global pixel repository, and maintaining temporal consistency in the generated videos can be framed as an optimization problem with a closed-form solution.
Then, the proposed approach enables high-quality and temporally consistent video-to-video translation by iteratively aligning corresponding noisy pixels across video frames using the provided temporal correspondence guidance derived from optical flows for each denoising step. Notably, MeDM does not necessitate additional video data for fine-tuning or test-time optimization of the DMs. With extensive quantitative, qualitative, and subjective evaluation on different benchmarks and tasks, our approach outperforms other state-of-the-art alternatives in terms of video quality and temporal consistency. The effectiveness of MeDM is also demonstrated in Figure \ref{fig:teaser}.

Our main contributions are summarized as follows.
\begin{itemize}
    \item We develop a practical coding algorithm and a global pixel repository based on optical flows to mediate independently estimated image scores. This mediation can seamlessly integrate into current state-of-the-art image-based denoising cycles, enhancing their ability for high-quality, consistent video-to-video translation.

    \item We propose a workaround for modifying observation-space scores in LDMs, which enables more general LDM applications. 
    
    \item We show that the proposed approach is effective in other real-world applications, yielding promising subjective evaluations. These applications include text-guided video editing and video anonymization.
\end{itemize}

\section{Related Work}
In this section, we briefly review the recent relevant works of text-guided video-to-video generation and the mediation of multiple diffusion generations.

\subsection{Diffusion-Based Video-to-Video Translation}

To address the resource-intensive nature of training video DMs, various zero-shot techniques have arisen for adapting existing image DMs to video-to-video translation. For instance, Pix2Video \cite{ceylan2023pix2video} employs a pre-trained structure-guided DM with self-attention features for modifying each denoising step. Rerender A Video \cite{yang2023rerender} ensures consistency in global style and texture using a hierarchical inter-frame restricted DM, with the generated key frames propagating to in-between frames through temporal-sensitive patch matching and blending. ControlVideo \cite{zhang2023controlvideo}, an extension of ControlNet \cite{zhang2023controlnet}, incorporates cross-frame interaction for visual consistency and video quality preservation. Additionally, Control-A-Video \cite{chen2023control-a-video} introduces motion-enhanced noise initialization and conditioning techniques for consistent and text-aligned videos, even with small datasets. In addition, there are other similar works \cite{xing2023make,wu2022tune-a-video,shin2023edit,liu2023video,qi2023fatezero,khachatryan2023text2video} and concurrent works \cite{ouyang2023codef,wu2023lamp,geyer2023tokenflow}. Although the aforementioned techniques succeed in transforming videos to different styles, artifacts and flickering in the video remains clearly visible. In comparison, our approach enables efficient and effective harmonious video editing.

\subsection{Mediating Multiple Diffusion Samples}
MultiDiffusion \cite{bar-tal2023multidiffusion} samples separate views from a DM. A view is a subset of pixels in the final image, which have intersection. The DM gives diverse score estimates for each view's intersection during denoising. \citeauthor{bar-tal2023multidiffusion} present view harmonization as a least-square optimization problem. In practice, they show a closed form solution as averaging views over the intersection during iterative denoising process, while testing mainly on simple views like rectangular crops and quantized masks from low-dimensional latent vectors of LDMs \cite{rombach2022LDM}. In contrast, our approach deals with more general views spanning temporal and spatial dimensions. These views can be subsets of consecutive and nonconsecutive pixels in video frames on the original dimension.

\begin{figure*}[t]
    \centering
    \includegraphics[width=0.8\linewidth]{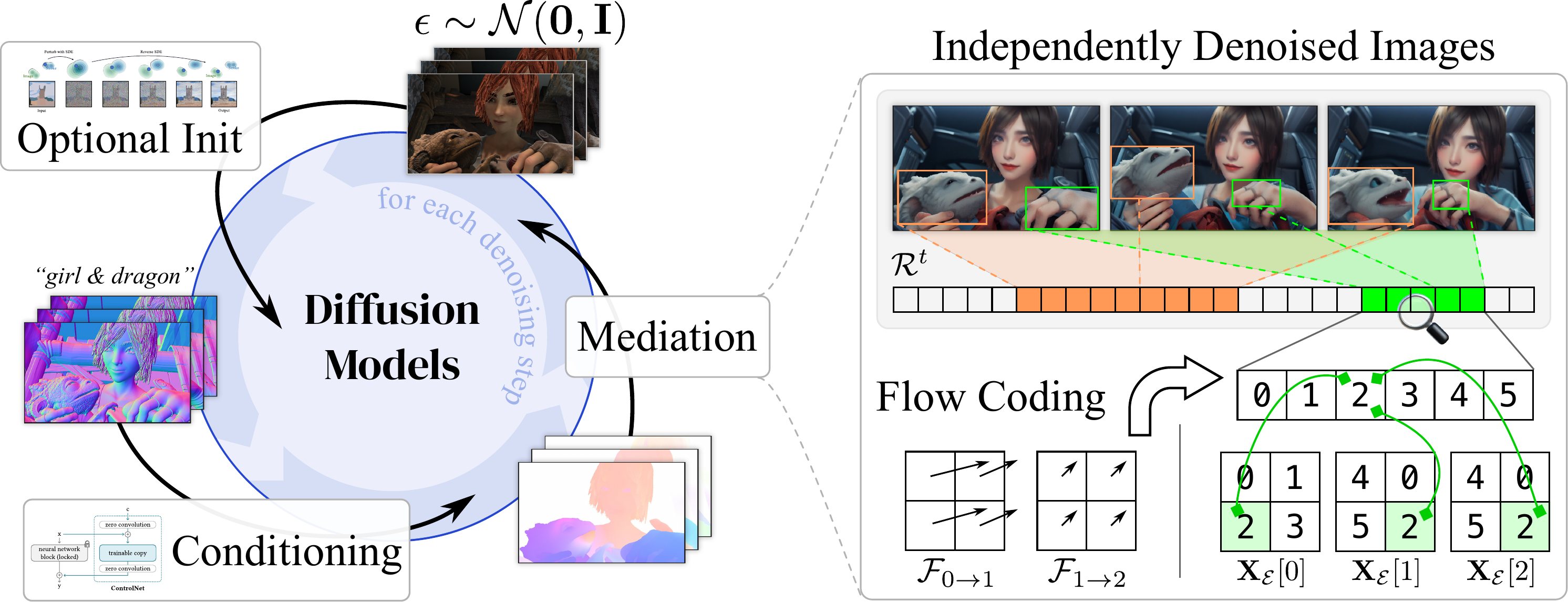}
    \caption{\textbf{Left: the proposed framework for video-to-video translation.} We extend the previous works in the image domain \cite{meng2022sdedit,zhang2023controlnet} to the video domain without any fine-tuning or iterative optimization. Our method mediates independent image score estimations after every denoising step, making them fluent motion pictures when viewed sequentially. Image credits to aforementioned authors. \textbf{Right: an illustration of a global pixel repository for a video and the proposed Flow Coding.} Video pixels are essentially views to the underlying objects. We construct an explicit pixel repository $\mathcal{R}^t$ to represent the underlying world. $\mathcal{R}^t$ is derived from the optical flows $\mathcal{F}$ through the proposed Flow Coding and stores all unique pixels of the video. The encoded frames $\mathbf{X}_\mathcal{E}$ and the repository $\mathcal{R}^t$  enable efficient harmonization of the divergent frame-wise score estimations during the generation process of Diffusion Models. Concretely, the bottom left pixel of $\mathbf{X}_{\mathcal{E}}[0]$, the bottom right pixels of $\mathbf{X}_{\mathcal{E}}[1]$ and $\mathbf{X}_{\mathcal{E}}[2]$ are all associated and synchronized with pixel \texttt{2} in $\mathcal{R}^t$.}
    \label{fig:system-diagram}
\end{figure*}

\section{Optical Flow-Based Temporal Correspondence}
In this section, we introduce Flow Coding, a physics-inspired coding at the core of the proposed method. Next, we propose an objective function with a closed-form solution for harmonization based on Flow Coding.

\subsection{Flow Coding}
\label{sec:flow-coding}

Beginning with the notion that the color\footnote{The shade, though subject to changes due to surrounding and environmental lighting, is intentionally disregarded in this study for the sake of brevity, with plans to address it in future research.} of a point on a real-world material remains constant over time when the material possesses a time-invariant electronic structure, we apply this concept to rasterized digital video. Within such videos, points are collectively perceived within a pixel via the sampling theorem. We extend this idea by assuming that pixels through which the same set of underlying analog points are viewed should exhibit identical colors. To achieve this, we use a pixel repository to represent all rasterized points of the observed object. Each video frame serves as a view of this pixel repository, with views being defined by pixel-level optical flows, as depicted in Figure \ref{fig:system-diagram}.

\begin{figure}[t]
    \centering
    \includegraphics[width=0.8\linewidth]{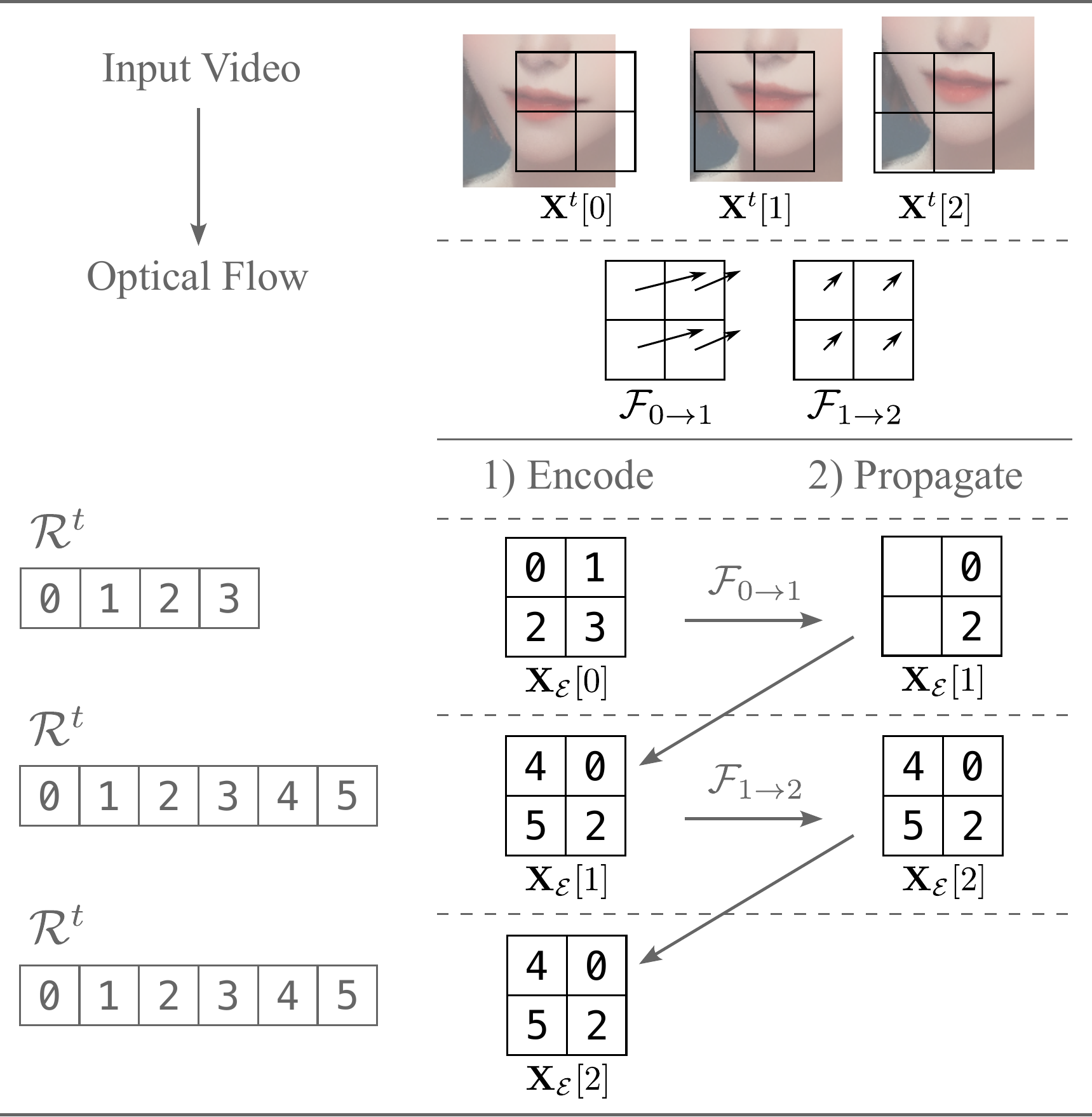}
    \caption{A toy example for Flow Coding. If the flow magnitude is larger than a pixel, \textit{e.g.} $\mathcal{F}_{0 \rightarrow 1}$, codes are propagated to the displaced position. If not, \textit{e.g.} $\mathcal{F}_{1 \rightarrow 2}$, codes are propagated to the same position.}
    \label{fig:flow-coding}
\end{figure}

In practice, we introduce Flow Coding, which uses per-frame optical flows and occlusions between adjacent frames. It encodes video pixels as indices pointing to a pixel repository $\mathcal{R}^t$. Flow Coding operates by iteratively performing two procedures through all frames: 1) Pixels in the currently processing frame are encoded and appended to $\mathcal{R}^t$ if they haven't been encoded before. 2) Codes from the encoded frame are propagated to corresponding pixels on the same motion trajectory in the next frame. For a concrete toy example of Flow Coding, refer to Figure \ref{fig:flow-coding}. We also provide a complete algorithm in Appendix \ref{app:flow-coding}.

\subsection{Global Harmonization}
\label{sec:global-harmonization}

After applying the Flow Coding algorithm, let $\mathbf{X}_{\mathcal{E}}$ represent the encoded frames (tensors of indices). To decode frames, we adopt the indexing notation $\mathcal{R}^t\left[\mathbf{X}_{\mathcal{E}}\right]$, which retrieves pixel values from the repository. Placing this notation on the right side of an equation fetches pixel values, while on the left side, it denotes references to store pixels.

During the denoising process of a DM, we let $\mathbf{X}^t$ be the noisy samples from the model evaluation of step $t$. The DM we consider here has no temporal information, and each video frame in $\mathbf{X}^t$ is generated independently. Therefore, to enforce temporal constraint, we construct a temporal consistency loss:
\begin{equation}
\label{eq:global-harmonization}
\mathcal{L}^t = \lVert \mathcal{R}^t\left[\mathbf{X}_{\mathcal{E}}\right] - \mathbf{X}^t \rVert_2,
\end{equation}
\begin{equation}
\mathcal{R}^t\left[\mathbf{X}_{\mathcal{E}}\right] \leftarrow G\left(\mathbf{X}^t\right),
\end{equation}
where
$\lVert \cdot \rVert_2$ denotes a tensor-norm taking square root over the square sum of all tensor entries,
$G$ is a function that mixes the pixels in $\mathbf{X}^t$ into the ones in $\mathcal{R}^t\left[\mathbf{X}_{\mathcal{E}}\right]$, and $\mathcal{R}^t$ is the pixel repository at time $t$. Notably, $\mathcal{R}^t\left[\mathbf{X}_{\mathcal{E}}\right]$ contains significant less unique pixels than $\mathbf{X}^t$, and $G$ is required to harmonize the associated pixels in $\mathbf{X}^t$ into a common values before they can be assigned to $\mathcal{R}^t\left[\mathbf{X}_{\mathcal{E}}\right]$.

The intuition behind Equation~\ref{eq:global-harmonization} is that under the physical constraint posed by the coding, we would like to find a frame composition that matches the unconstrained $\mathbf{X}^t$ as close as possible. To solve the system analytically, we use a post-hoc
solver following \cite{bar-tal2023multidiffusion}
\begin{equation}
\label{eq:global-harmonization-solver}
G_{avg} = \operatorname*{arg\,min}_G\mathcal{L}^t,
\end{equation}
where $G_{avg}$ performs unweighted averages on pixels in $\mathbf{X}^t$ that are associated with the same pixel in $\mathcal{R}^t$ according to $\mathbf{X}_{\mathcal{E}}$. In practice, $G_{avg}$ can be implemented in Pytorch in three lines of code (Listing~\ref{lst:unweighted-global-harmonization}).

\begin{listing}[tb]%
\caption{Unweighted global harmonization in Pytorch}%
\label{lst:unweighted-global-harmonization}%
\begin{lstlisting}[language=Python]
# assume accumulate=True
cnt.index_put_((X_E,), 1)
repo.index_put_((X_E,), X_t)
avg = torch.where(cnt>0, repo/cnt, repo)
\end{lstlisting}
\end{listing}

\section{Mediating Image Diffusion Samples with Temporal Correspondence Guidance}
In this section, we briefly review the essential formulation in DMs and introduce the proposed temporal correspondence guidance along with techniques to make it compatible with LDMs.
\subsection{Diffusion Models}
Given a predefined noise scheduling $\{\alpha^t | t \in [0, T]\}$ and a data point $\mathbf{X}^0$, the noisy sample $\mathbf{X}^t$ for step $t$ is defined as
\begin{equation}
\label{eq:diffusion-x_t}
\mathbf{X}^t = \sqrt{\alpha^t} \mathbf{X}^0 + \sqrt{1-\alpha^t} \boldsymbol{\epsilon}^t, \; \text{where} \; \boldsymbol{\epsilon}^t \sim \mathcal{N}(\mathbf{0}, \mathbf{I}). \\
\end{equation}
Equivalently,
\begin{align}
\label{eq:diffusion-x_0}
\mathbf{X}^0 &= \frac{1}{\sqrt{\alpha^t}} \left(\mathbf{X}^t - \sqrt{1-\alpha^t} \boldsymbol{\epsilon}^t \right), \\
\label{eq:diffusion-eps}
\boldsymbol{\epsilon}^t &= \frac{1}{\sqrt{1-\alpha^t}} \left( \mathbf{X}^t - \sqrt{\alpha^t} \mathbf{X}^0 \right).
\end{align}
A neural network parameterized by $\theta$ is often used to predict either data reconstruction $\mathbf{X}^0_\theta(\mathbf{X}^t)$ or noise residual $\boldsymbol{\epsilon}^t_\theta(\mathbf{X}^t)$ for the construction of a less noisy data distribution.

\subsection{Temporal Correspondence Guidance}
In this section, we introduce how to leverage the pixel repository $\mathcal{R}^t$ to help an image DM generate temporally consistent video frames. For each model evaluation of the DM parameterized by $\theta$, we modify the noisy samples as follows:
\begin{equation}
\label{eq:harmonization-guidance-x_t}
\mathbf{X}^t_\theta \leftarrow (1 - w) \mathbf{X}^t_\theta +  w \mathcal{R}^t\left[\mathbf{X}_{\mathcal{E}}\right],
\end{equation}
or using the score estimation following \citeauthor{ho2021cfg}:
\begin{equation}
\label{eq:harmonization-guidance-eps}
\boldsymbol{\epsilon}^t_\theta \leftarrow (1 - w) \boldsymbol{\epsilon}^t_\theta +  w \mathcal{R}^t_{\boldsymbol{\epsilon}}\left[\mathbf{X}_{\mathcal{E}}\right],
\end{equation}
\begin{equation}
\label{eq:R_eps-wrong}
\mathcal{R}^t_{\boldsymbol{\epsilon}}\left[\mathbf{X}_{\mathcal{E}}\right] \leftarrow G\left(\boldsymbol{\epsilon}^t_\theta\right),
\end{equation}
with weight $w$ on the harmonized samples that controls the strength of temporal correspondence guidance\footnote{It is possible to use the extrapolation formulation, but we found this lead to serious early saturation in our experiments.} and $\mathcal{R}^t_\epsilon$ for scores is analogous to $\mathcal{R}^t$ for noisy pixels. However, Equation~\ref{eq:R_eps-wrong} does not support LDMs \cite{rombach2022LDM} because they perform the denoising process in a spatially smaller dimension, and $G$ only operates on the original dimension. Moreover, their perceptual Autoencoders work with neither noisy samples $\mathbf{X}^t$ nor noise predictions $\boldsymbol{\epsilon}^t$, so it is infeasible to cast the samples to a proper space when needed without further fine-tuning the Autoencoders.

To address this issue, we leverage Equation~\ref{eq:diffusion-x_0} to derive an inaccurate prediction of the original samples $\hat{\mathbf{X}}^{0, t}$ from the predicted noise $\boldsymbol{\epsilon}^t_\theta$. Empirically, we find that $\hat{\mathbf{X}}^{0, t}$ can be faithfully cast between latent space and observed space by pre-trained Autoencoders even in the early stage of denoising process. Therefore, incorporating Equation~\ref{eq:diffusion-eps}, we can replace Equation~\ref{eq:R_eps-wrong} with
\begin{equation}
\label{eq:R_eps-correct}
\mathcal{R}^t_{\boldsymbol{\epsilon}}\left[\mathbf{X}_{\mathcal{E}}\right] \leftarrow \frac{1}{\sqrt{1-\alpha_t}} \left( \mathbf{X}_t - \sqrt{\alpha_t} \Phi_e \left(G\left(\Phi_d\left(\hat{\mathbf{X}}^{0, t}\right)\right)\right) \right)
\end{equation}
to provide compatibility of LDMs, where $\Phi_e$ and $\Phi_d$ are the encoder and the decoder of the Autoencoder, respectively.

\subsection{Enhancing Video Sharpness}
In our early experiments, we found that harmonizing independent score estimations from standard DMs directly results in suboptimal blurry video due to significant visual structure disparities between optical flows and DM estimates. To address this, we incorporate SDEdit \cite{meng2022sdedit} and ControlNet \cite{zhang2023controlnet} into our approach. These methods integrate structural and semantic priors, with ControlNet excelling in fusing input structural information and pre-trained DM semantic priors. 

\section{Experiments}
In this section we show that our method outperform other baseline methods in various benchmarks and applications, supported by extensive quantitative metrics, qualitative comparison and human evaluation.

\paragraph{Setup}
For all video generation, we use DDIM scheduler \cite{song2021ddim} with 20 denoising steps and set the temporal correspondence guidance scale $w$ to $0.8$. We use Stable Diffusion \cite{rombach2022LDM} with the checkpoint from Civitai\footnote{\url{https://civitai.com/models/43331/majicmix-realistic}} as our pre-trained DM to match up with the baseline methods. For all estimated optical flows, we use UniMatch \cite{xu2023unimatch}, and we use Informative Drawings \cite{chan2022lineart} for all lineart translations which serve as the input conditions to ControlNet unless further specified.\footnote{Our method also works with more recent SDXL in theory. However, the ControlNet adapters of certain modalities for SDXL are not available yet at the time of our project development.} We do not provide a dedicated prompt to each video generation (except for text-guided video editing). Instead, reasonable texture and colors are automatically inferred through our cross-frame harmonization together with ControlNet's guess mode. Finally, We conduct online user studies for all tasks in this section. The feedback is collected from a total of 71 people with diverse background. For most of the questions, they are ask to give a score ranging from 1 (the worst) to 5 (the best).

\subsection{Video Rendering}
Our method is capable of efficiently rendering high quality videos solely from 3D assets. We run the experiments on MPI Sintel \cite{butler2012sintel} and Virtual KITTI 2 \cite{cabon2020vkitti2}, synthetic video datasets that provides ground-truth (GT) 3D assets, including optical flows, occlusions, position information (depth, normal) and rendered animations. We consider two variants to this task.

\paragraph{(Full) Rendering.} All methods generate videos solely from optical flows and position information (we consider normal maps here). The result video should differ greatly from the original rendered animation in terms of surface texture and colors. We compare our framework with 4 baseline methods: ControlNet \cite{zhang2023controlnet}, ControlVideo \cite{zhang2023controlvideo}, Video ControlNet \cite{chu2023video} and Rerender A Video \cite{yang2023rerender}.

\paragraph{Assistive Rendering.} 
The methods capable of this task begin with adding noise to the pre-rendered videos and perform the denoising process from $0.5T$ step following SDEdit \cite{meng2022sdedit}. The generated video should be similar to the original animation and incorporate the realistic prior from the pre-trained DM.

\begin{figure*}
\centering
\begin{tabular}{lm{22.8mm}m{22.8mm}m{22.8mm}m{22.8mm}m{22.8mm}l}
& First frame & Horizontal scan & First frame & Horizontal scan & First frame & Horizontal scan \\
(a) & \multicolumn{6}{c}{\includegraphics[align=c,width=0.9\linewidth]{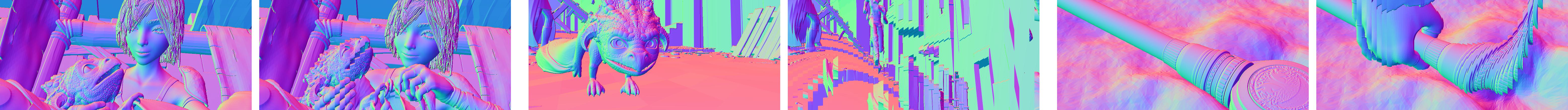}} \\ [5mm]
(b) & \multicolumn{6}{c}{\includegraphics[align=c,width=0.9\linewidth]{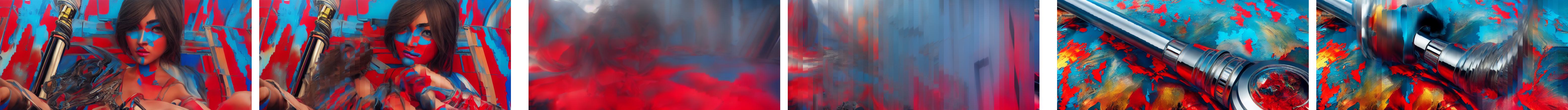}} \\ [5mm]
(c) & \multicolumn{6}{c}{\includegraphics[align=c,width=0.9\linewidth]{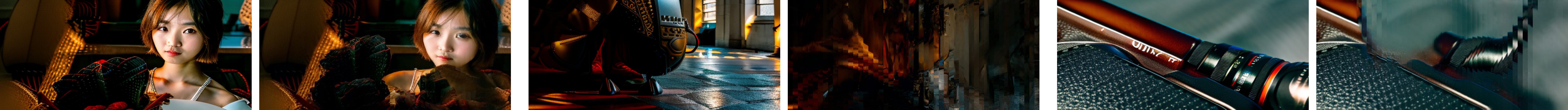}} \\ [5mm]
\textbf{(d)} & \multicolumn{6}{c}{\includegraphics[align=c,width=0.9\linewidth]{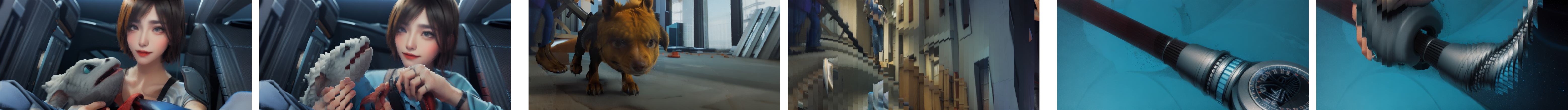}} \\ [4mm]
\midrule
(e) & \multicolumn{6}{c}{\includegraphics[align=c,width=0.9\linewidth]{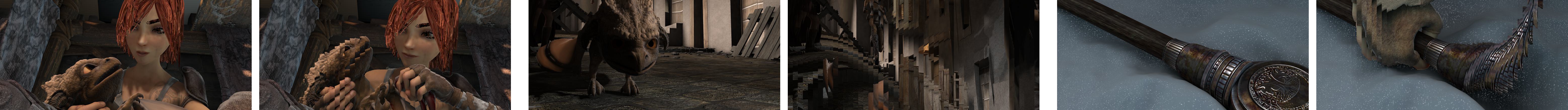}} \\ [5mm]
(f) & \multicolumn{6}{c}{\includegraphics[align=c,width=0.9\linewidth]{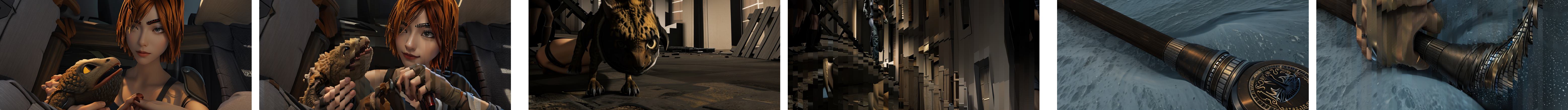}} \\ [5mm]
(g) & \multicolumn{6}{c}{\includegraphics[align=c,width=0.9\linewidth]{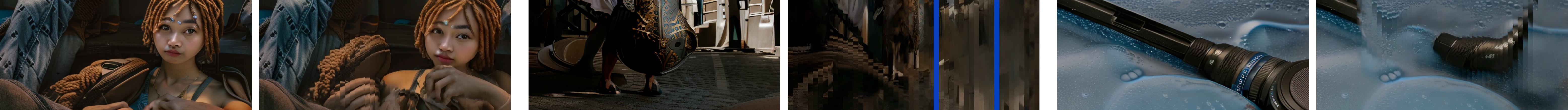}} \\ [5mm]
\textbf{(h)} & \multicolumn{6}{c}{\includegraphics[align=c,width=0.9\linewidth]{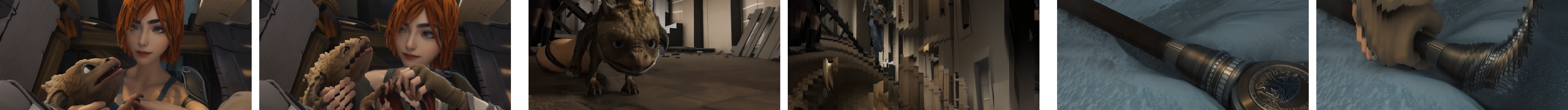}} \\
\end{tabular}
\caption{\textbf{Visual comparison for video rendering (a-d) and assistive rendering (e-h).} For each method, we provide the first frame and the horizontal scan for visual quality and temporal coherence comparison, respectively. The horizontal scan is composed of the left-shifting vertical segments of pixels from each video frame. A fluent video should reconstruct a stripe-free scan in static area (e.g. background, see the GT video (a,e)). From top to bottom, the presented videos are from (a) input GT normals, (b) ControlVideo, (c) Rerender A Video, \textbf{(d) ours (estimated flow)}, (e) input GT animations, (f) ControlNet, (g) Rerender A Video and \textbf{(e)  ours (estimated flow)}.}
\label{fig:rendering}
\end{figure*}

For our method, we use the lineart derived from the normal maps as the input conditions to ControlNet\footnote{Although the pre-trained ControlNet for the depth and normal maps are available, the depth maps fail to encode subtle details like facial features, and the model for normal maps is trained on the estimated normal maps which differ greatly from the GT ones}. The qualitative comparisons in Figure \ref{fig:rendering} show that our generated videos are high-quality, fluent, and faithfully follow the input conditioning for full video rendering. For assistive rendering, our method further enhances the realism of objects with complex texture, such as hair.

\begin{figure}[t]
\centering
\begin{subfigure}{0.49\linewidth}
\begin{tikzpicture}
\pgfplotsset{every tick label/.append style={font=\tiny}}
\pgfplotstableread[col sep=comma]{guidance-scale.csv}\datatable
\begin{axis}[
xlabel={$w$},
xticklabel={\pgfmathprintnumber[assume math mode=true]{\tick}},
yticklabel={\pgfmathprintnumber[assume math mode=true]{\tick}},
width=5cm, 
height=4cm 
]
\addplot table[x={w}, y={epe}] {\datatable};
\end{axis}
\end{tikzpicture}
\caption{End-point-error (EPE)}
\label{fig:guidance-scale-epe}
\end{subfigure}
\begin{subfigure}{0.49\linewidth}
\begin{tikzpicture}
\pgfplotsset{every tick label/.append style={font=\tiny}}
\pgfplotstableread[col sep=comma]{guidance-scale.csv}\datatable
\begin{axis}[
xlabel={$w$},
xticklabel={\pgfmathprintnumber[assume math mode=true]{\tick}},
yticklabel={\pgfmathprintnumber[assume math mode=true]{\tick}},
width=5cm, 
height=4cm 
]
\addplot table[x={w}, y={>1}] {\datatable};
\addlegendentry{\footnotesize 1px}
\addplot table[x={w}, y={>3}] {\datatable};
\addlegendentry{\footnotesize 3px}
\addplot table[x={w}, y={>5}] {\datatable};
\addlegendentry{\footnotesize 5px}
\end{axis}
\end{tikzpicture}
\caption{Portion of errors larger than}
\label{fig:guidance-scale-epe-portion}
\end{subfigure}
\caption{\textbf{EPE curves over the temporal correspondence guidance strengths.} The videos are generated from pure noise and conditioned on the assets from MPI Sintel. ($w = 0$ refers to the vanilla ControlNet).}
\label{fig:guidance-scale}
\end{figure}
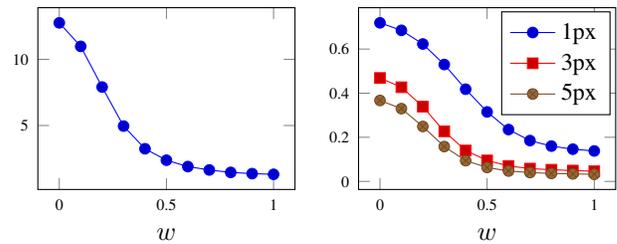

We also provide an objective and automatic quantitative measure for the video fluency. We estimate the optical flows from the generated videos and compare the flows with the GT flows. We use the average endpoint error (EPE) between the flows to assess the temporal consistency in videos. Figure \ref{fig:guidance-scale} validates that a fluent video leads to a lower EPE, and vice versa. As shown in Table \ref{tab:sintel-vkitti2}, our method has the lowest EPEs among all methods in each task, and it is the closest to the one of the animation on which the flow estimator is trained.

Finally, we close this task with a subjective test. We ask the participants how the presented videos are in terms of quality, sharpness, and fluency. We further ask them to rate the realism of the videos, and the results in Table \ref{tab:render-user} shows that in addition to having the highest quality, our videos are even more realistic than the animation videos from the dataset.

\begin{table}[t]
\centering
\begin{tabular}{llcc}
    \toprule
    & Method & Rendering & Assist. Rendering \\
    \midrule
    \multirow{8}{*}{\rotatebox{90}{\small MPI Sintel}}
    & ControlNet & 12.757 & 5.924 \\
    & ControlVideo & 12.757 & N/A \\
    & Video ControlNet & 12.878 & N/A \\
    & Rerender A Video & 7.953 & 7.775 \\
    & Ours (Est. flow) & 1.501 & 1.570 \\
    & Ours (GT flow) & \textbf{1.456} & \textbf{1.202} \\
    \cmidrule(){2-4}
    & Animation & \multicolumn{2}{c}{0.403} \\

    \midrule
    
    \multirow{5}{*}{\rotatebox{90}{\small VKITTI 2}}
    & ControlNet & 12.575 & 5.070 \\
    & Video ControlNet & 15.285 & N/A \\
    & Ours (Est. flow) & 2.857 & 2.695 \\
    & Ours (GT flow) & \textbf{2.483} & \textbf{2.217} \\
    \cmidrule(){2-4}
    & Animation & \multicolumn{2}{c}{1.737} \\

    \bottomrule
\end{tabular}
\caption{\textbf{EPE for video rendering on MPI Sintel and Virtual KITTI 2.} The best methods are in \textbf{bold}.}
\label{tab:sintel-vkitti2}
\end{table}

\begin{figure*}
\centering
\begin{tabular}{lm{16mm}m{16mm}m{16mm}m{16mm}m{16mm}m{16mm}m{16mm}l}
& First frame & Scan & First frame & Scan & First frame & Scan & First frame & Scan \\
(a) & \multicolumn{8}{c}{\includegraphics[align=c,width=0.9\linewidth]{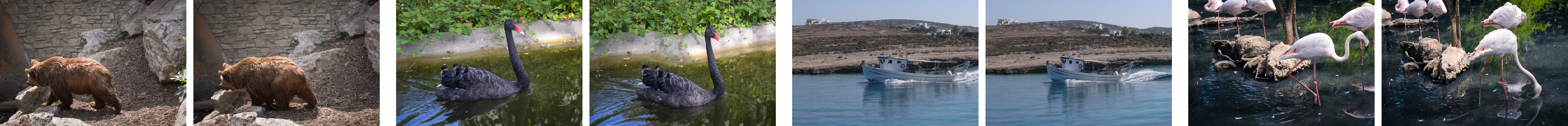}} \\ [5mm]
(b) & \multicolumn{8}{c}{\includegraphics[align=c,width=0.9\linewidth]{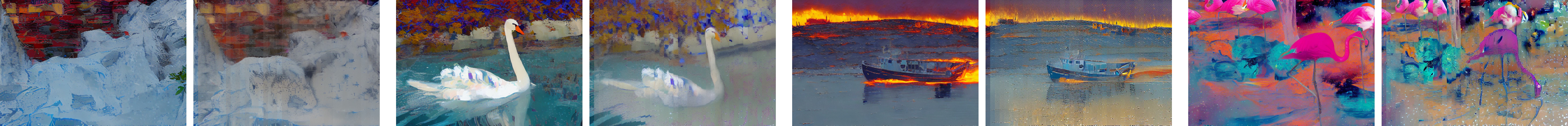}} \\ [5mm]
(c) & \multicolumn{8}{c}{\includegraphics[align=c,width=0.9\linewidth]{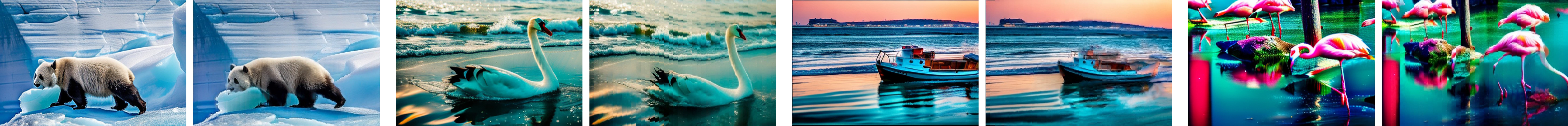}} \\ [5mm]
\textbf{(d)} & \multicolumn{8}{c}{\includegraphics[align=c,width=0.9\linewidth]{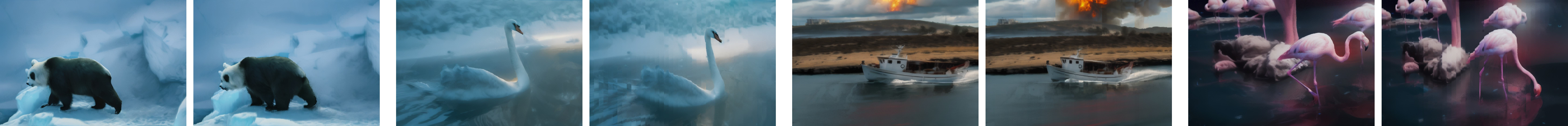}} \\ [5mm]
\textbf{(e)} & \multicolumn{8}{c}{\includegraphics[align=c,width=0.9\linewidth]{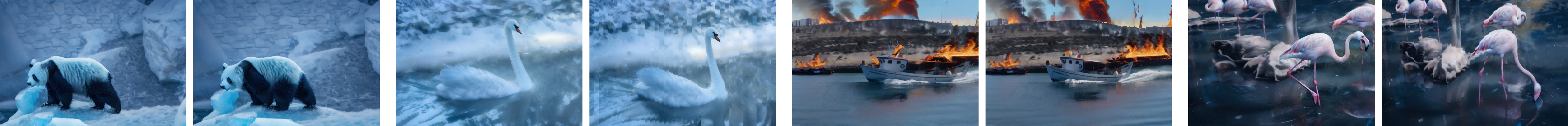}} \\
& \multicolumn{2}{m{32mm}}{\textit{a panda in Arctic, snow, ice, iceberg}} & \multicolumn{2}{m{32mm}}{\textit{a white swan in the ocean, snowing}} & \multicolumn{2}{m{32mm}}{\textit{a boat on fire}} & \multicolumn{2}{c}{\textit{flamingos in outer space}} \\
\end{tabular}
\caption{\textbf{Visual comparison for text-guided video editing.} For each method, we provide the first frame and the horizontal scan following Figure \ref{fig:rendering}. From top to bottom, the presented videos are from (a) input video, (b) Control-A-Video, (c) Rerender A Video, \textbf{(d) ours (Lineart)} and \textbf{(d) ours (InstructPix2Pix)}.}
\label{fig:edit}
\end{figure*}

\begin{table}[tb]
\centering
\begin{tabular}{llcc}
\toprule
\multicolumn{2}{l}{Method} & Video quality & Realism  \\
\midrule
\multirow{6}{*}{\rotatebox{90}{\small Non-Assistive}}

& ControlNet & 1.662 & 2.718 \\
& ControlVideo & 2.437 & 1.775 \\
& Control-A-Video & 3.662 & 1.606 \\
& Video ControlNet & 2.197 & 2.380 \\
& Rerender A Video & 2.183 & 2.127 \\
& Ours & \textbf{4.423} & \textbf{4.014} \\
\midrule
& Animation & 4.423 & 3.113 \\
\midrule
\multirow{3}{*}{\rotatebox{90}{\small Assistive}}
& ControlNet & 2.465 & 2.408 \\
& Rerender A Video & 1.887 & 2.479 \\
& Ours & \textbf{4.380} & \textbf{3.958} \\

\bottomrule
\end{tabular}
\caption{\textbf{Human evaluation for video rendering on Sintel.}}
\label{tab:render-user}
\end{table}

\subsection{Video-to-Video Translation}
To show that our method perform well even without high precision optical flows, we conduct evaluation on real videos. Considering two applications of video-to-video translation, we perform text-guided video editing on the videos in DAVIS 2016 \cite{perazzi2016davis} and video anonymization on the videos from CelebV-HQ \cite{zhu2022celebvhq}.

\paragraph{Text-Guided Video Editing.}
We take videos from DAVIS 2016 and caption them with seemingly absurd descriptions to test the models' ability in combining conflicting ideas into the same canvas. In addition to using the lineart for ContolNet, we also test the InstructPix2Pix version of ControlNet, which directly consumes real videos. We further consider an inversion-based approach, Pix2Video \cite{ceylan2023pix2video}, for this task and provide visual comparisons in Figure \ref{fig:edit}. Together with the subjective scores in Table \ref{tab:edit}, our methods, especially for the InstructPix2Pix variant, produce the best video quality and image-text alignment.

\begin{table}[tb]
\centering
\begin{tabular}{lcc}
\toprule
Method & Video quality & Text alignment \\
\midrule

ControlNet & 2.377 & 3.289 \\
Pix2Video & 1.451 & 1.592 \\
ControlVideo & 2.289 & 2.430 \\
Control-A-Video & 2.634 & 1.859 \\
Rerender A Video & 3.042 & 3.099 \\
Ours (Lineart) & \textbf{4.042} & 3.810 \\
Ours (Instruct P2P) & 3.901 & \textbf{4.338} \\

\bottomrule
\end{tabular}
\caption{\textbf{Human evaluation for text-guided video editing on DAVIS 2016.} The best methods are in \textbf{bold}.}
\label{tab:edit}
\end{table}

\begin{table}[tb]
\centering
\begin{tabular*}{\linewidth}{@{\extracolsep{\fill}} l|ccc}
\toprule
Method & {\small Recognizability} & {\small Realism} & {\small Faithfulness} \\
\midrule

DeepPrivacy & 63.01\% & 2.019 & 4.216 \\
Ours & \textbf{20.83\%} & \textbf{3.507} & \textbf{4.258} \\

\bottomrule
\end{tabular*}
\includegraphics[width=\linewidth]{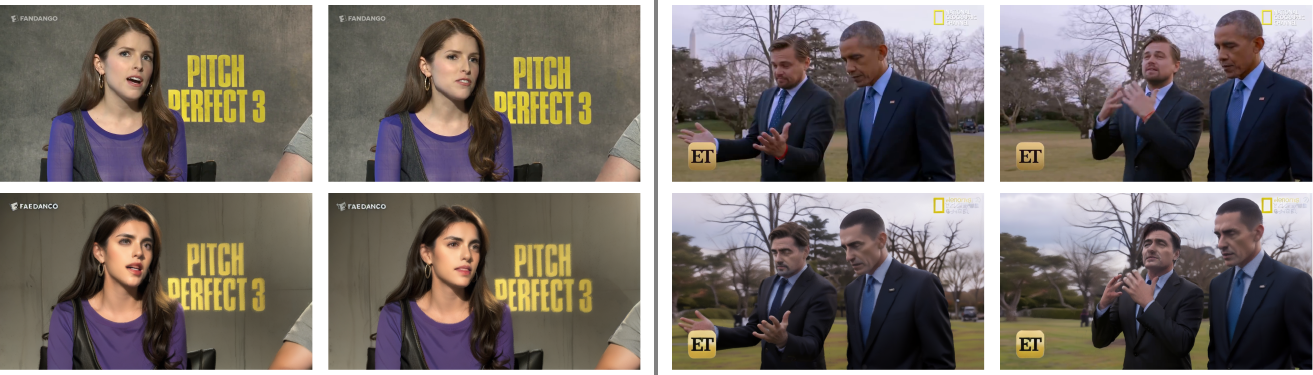}
\caption{\textbf{User feedback on video anonymization.} \textbf{Recognizability:} the proportion of participants correctly identifying the celebrity. \textbf{Realism:} how realistic is the anonymized video. \textbf{Faithfulness:} how much content except the identity is preserved. \textbf{Upper row images:} the original clips. \textbf{Lower row images:} the anonymized clips (ours only).}
\label{tab:anonymization}
\end{table}

\begin{table}[tb]
\centering
\begin{tabular*}{\linewidth}{@{\extracolsep{\fill}} cc|c|l}
    \toprule
    SDEdit & ControlNet & EPE & Comment \\
    \midrule
    \ding{55}& \ding{55} & 3.781 & Prompted \\
    \ding{51}& \ding{55} & 1.798 & - \\
    \ding{55}& \ding{51} & 1.456 & {\small Rendering} \\
    \ding{51}& \ding{51} & 1.202 & {\small Assistive rendering} \\
    \bottomrule
\end{tabular*}
\begin{tabular*}{\linewidth}{@{\extracolsep{\fill}} c|cc}
& \ding{55} ControlNet & \ding{51} ControlNet \\
\midrule
\rotatebox{90}{\ding{55} SDEdit} & \includegraphics[width=0.4\linewidth]{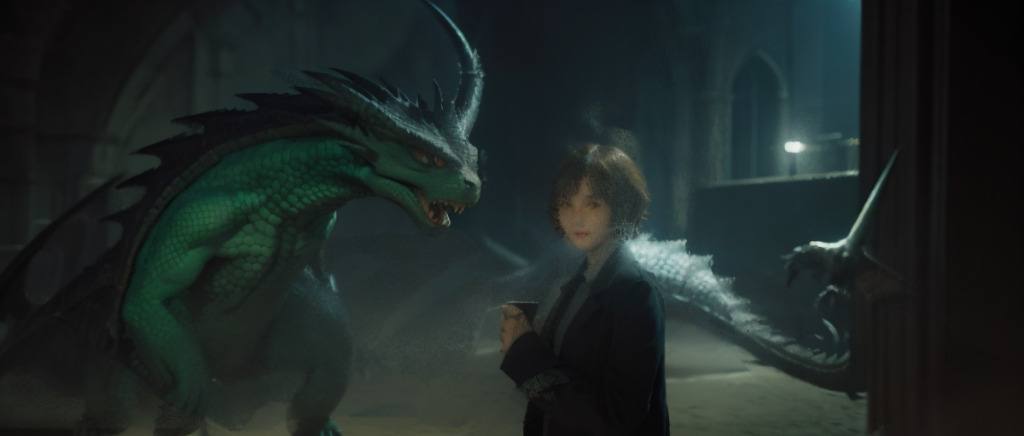} &
\includegraphics[width=0.4\linewidth]{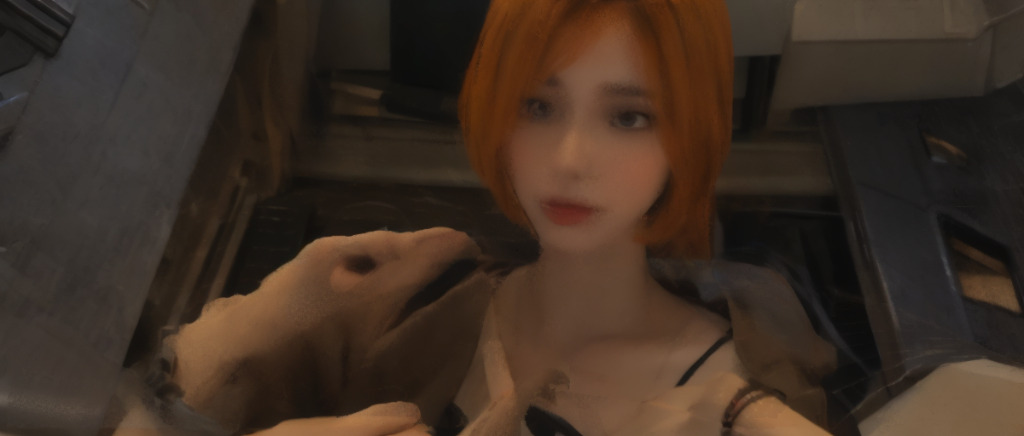} \\
\rotatebox{90}{\ding{51} SDEdit} & \includegraphics[width=0.4\linewidth]{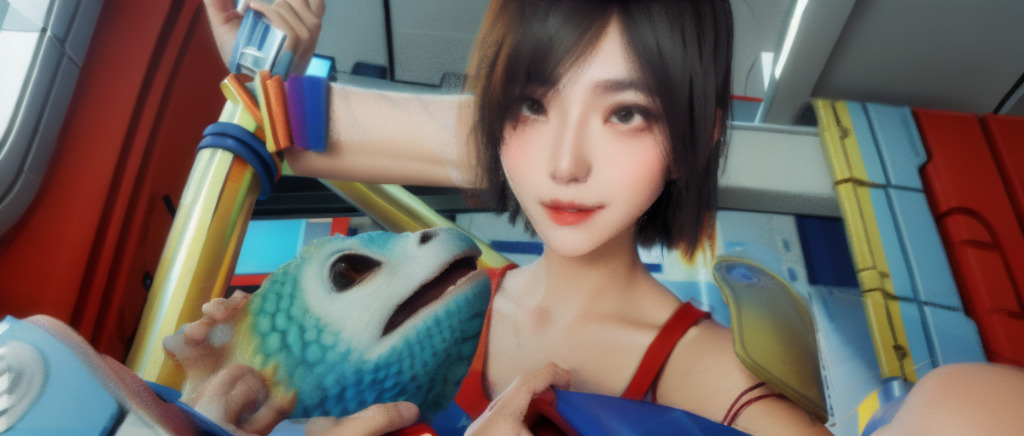} &
\includegraphics[width=0.4\linewidth]{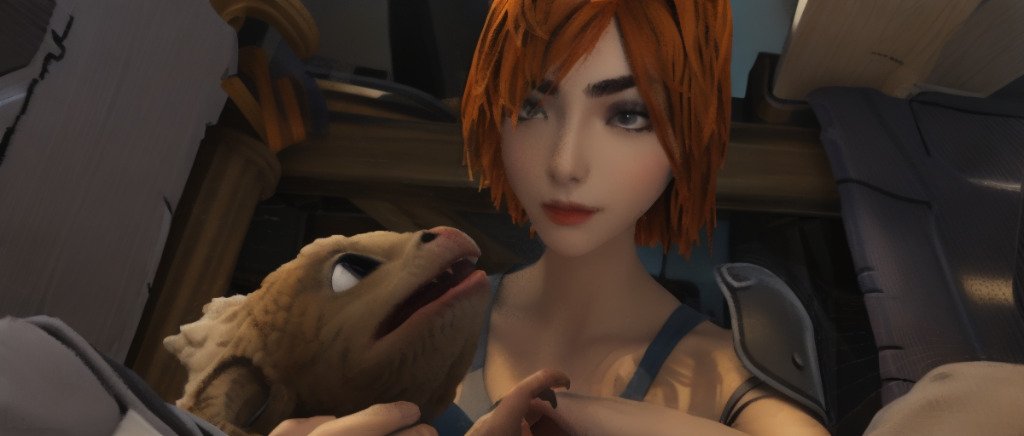} \\
\end{tabular*}
\caption{\textbf{Ablation study for ControlNet and SDEdit.}}
\label{tab:ablation-cn-sde}
\end{table}

\paragraph{Video Anonymization.} Anonymizing videos allows ones to analyze behaviors in videos without exposing individuals' identities. In order to make the anonymized video useful, we need to preserve everything in the video except the facial identity. We leverage the fact that human visual perception exhibits a remarkable sensitivity to human faces while our ability to detect and recognize other objects is not as specialized. We add noise to a video with a strength of $0.5T$, which is strong enough to erase the identity while preserving other objects and the background scene, and perform denoising using the proposed method to obtain the anonymized video. We conduct a user study to validate the effectiveness of our method. We compare with DeepPrivacy \cite{hukkelas2019deepprivacy} and report the results in Table \ref{tab:anonymization}. We show that a larger portion of the participants\footnote{We only consider the participants who know of the celebrities in advance.} cannot recognize the celebrities from our anonymized videos compared to the ones from DeepPrivacy. The participants also consider our method produce samples that are more realistic and preserves more details in the original videos.

\subsection{Without ControlNet or SDEdit}
Since we heavily utilize ControlNet and SDEdit in our experiments, we provide an ablation study in Table \ref{tab:ablation-cn-sde} to justify this usage. From both qualitative and quantitative results, we show that removing either one of the two increases the EPE. When using neither ControlNet nor SDEdit, our method fails to faithfully follow the visual structure hinted by the motion in optical flows.

\section{Conclusion and Future Work}
In this work, we propose a novel approach, MeDM, which employs optical flows to guide temporal correspondence. This methodology allows for efficient and high-quality video-to-video translation while preserving temporal consistency. By aligning pixel correspondences across video frames using optical flow-derived guidance, MeDM avoids the need for additional video data and achieves superior results on multiple benchmarks compared to existing methods.

While we leverage discretized optical flows for adjacent frames in this work, we are also interested in the potential of point tracking methods \cite{harley2022pips,doersch2023tapir,wang2023omnimotion}, which track pixels across occlusions and have continuous coordinate system. This helps the proposed Flow Coding build more accurate pixel correspondences. Despite their advantages, these approaches are currently costlier and less robust. Another potential enhancement for our method involves enabling structural translation. While optical flows incorporate structural details, they might not align well with objects that undergo significant contour changes. Finally, our method inherits the biases of the underlying Diffusion Models. However, it works with a large volume of community-driven LoRA/finetuned checkpoints, which may help to mitigate these biases.

\section{Acknowledgments}
This research is supported by National Science and Technology Council, Taiwan (R.O.C), under the grant number of NSTC-112-2634-F-002-006, NSTC-112-2222-E-001-001-MY2, and NSTC-110-2221-E-001-009-MY2, and Academia Sinica under the grant number of AS-CDA-110-M09.

\bibliography{aaai24}
\clearpage
\appendix
\section{System}
We run all experiments on the system with the specs listed below.

\vspace{0.2cm}
\noindent\begin{tabular}{l|l}
\toprule
CPU & Intel(R) Xeon(R) Silver 4210 CPU @ 2.20GHz \\
GPU & NVIDIA RTX A5000 24GB \\
OS & Ubuntu 20.04.6 LTS \\
\bottomrule
\end{tabular}
\vspace{0.2cm}

We heavily utilize the codebase of Hugging Face Diffusers\footnote{\url{https://huggingface.co/docs/diffusers/index}} in this work. The ControlNet weights for lineart\footnote{\url{https://huggingface.co/lllyasviel/control_v11p_sd15_lineart}} and InstructPix2Pix\footnote{\url{https://huggingface.co/lllyasviel/control_v11e_sd15_ip2p}} are also hosted on Hugging Face.

\section{Flow Coding}
\label{app:flow-coding}
To complement the toy example for Flow Coding in Figure \ref{fig:flow-coding}, we provide a complete algorithm to elaborate the process in Algroithm \ref{alg:flow-coding}.

\section{Reducing Discretization Errors}
Per-pixel optical flows are utilized in our study. These flows require rounding during Flow Coding to obtain the integer coordinates of the destination, leading to errors in code propagation due to discretization. To address this problem, we introduce a technique to mitigate these errors. \footnote{It is also possible to leverage point tracking methods \cite{harley2022pips}, but we leave it for future work.}
\paragraph{Optical flow for distant frames.}
When optical flows from distant frames are available alongside those from adjacent frames, these distant flows can be employed to counteract accumulated errors stemming from discretization. Specifically, when $d$ denotes the number of frames preceding the current one, the encoded frame $\mathcal{E}_{i+1}$ prioritizes codes from the distant frame $\mathcal{E}_{i-d}$ over the previous frame $\mathcal{E}_{i}$, such as
\begin{equation}
\label{eq:long-flow}
\mathcal{E}_{i+1}[\mathcal{V}] \leftarrow \mathcal{E}_{i-d}[\mathcal{D}_{i-d} \cap \mathcal{O}_{i-d}' \cap \mathcal{O}_{i}'] \cup \mathcal{E}_{i}[\mathcal{D}_{i} \cap \mathcal{O}_{i-d} \cap \mathcal{O}_{i}'],
\end{equation}
where $\mathcal{V}, \mathcal{D}$ and $\mathcal{O}$ are the coordinate vectors, destinations and occlusion, respectively, in line with Algorithm~\ref{alg:flow-coding}. The index ${i-d}$ is clamped to be non-negative. Nevertheless, our empirical observations suggest that distant optical flows tend to exhibit greater estimation errors, implying that their inclusion might not necessarily lead to improved performance

\begin{algorithm}[t]
\caption{Flow Coding}
\label{alg:flow-coding}
\textbf{Input}: Flows $\mathcal{F}$ and occlusions $\mathcal{O}$
\textbf{Output}: Encoded frames $\mathbf{X}_{\mathcal{E}}$ and number of unique pixels $n$ \\
\begin{algorithmic}[1] 
\IF {$\mathcal{F}$ is forward flow}
\STATE Temporally flip $\mathcal{F}$ and $\mathcal{O}$
\ENDIF
\STATE Let $n = height(\mathcal{F}) \times width(\mathcal{F})$
\STATE Let encoded frame $\mathcal{E}$ be a matrix with shape of $\mathcal{F}$
\STATE Fill $\mathcal{E}$ with codes from $0$ to $n-1$
\STATE Let $\mathbf{X}_{\mathcal{E}} = \{\mathcal{E}\}$
\FOR{each flow $\mathcal{F}_i$ and occlusion $\mathcal{O}_i$ in $\mathcal{F}$ and $\mathcal{O}$}
\STATE Let $\mathcal{E}_{prev} = \mathcal{E}$
\STATE Let $\mathcal{V}$ be coordinate vectors that cover every point in the spatial dimension of $\mathcal{F}_i$
\STATE Let warped destination $\mathcal{D} = \textbf{round}(\mathcal{V} + \mathcal{F}_i)$
\STATE Propagate codes: $\mathcal{E}[\mathcal{V}] = \mathcal{E}_{prev}[\mathcal{D}]$
\STATE Let coordinate vectors of novel pixel $\mathcal{N} = \mathcal{V}[\mathcal{O}]$
\STATE Fill $\mathcal{E}[\mathcal{V}]$ with codes from $n$ to $n + |\mathcal{N}| - 1$
\STATE $n = n + |\mathcal{N}|$
\STATE $\mathbf{X}_{\mathcal{E}} = \mathbf{X}_{\mathcal{E}} \cup \{\mathcal{E}\}$
\ENDFOR
\IF {$\mathcal{F}$ is forward flow}
\STATE Temporally flip $\mathbf{X}_{\mathcal{E}}$
\ENDIF
\RETURN $\mathbf{X}_{\mathcal{E}}$, $n$
\end{algorithmic}
\end{algorithm}



\section{Local Harmonization}
\label{sec:local-harmonization}
Global harmonization represents a rigorous process, enforcing consistent pixel values over time to visualize a consistent physical point. This approach, however, disregards alterations in shading due to varying lighting conditions. In this section, we introduce a preliminary solution that offers a more relaxed approach to this procedure. Drawing inspiration from the notion that distant pixels are likely to exhibit greater discrepancies than adjacent ones, we organize the pixels observing the same object point into a 1D tensor according to their temporal sequence. To enhance coherence, we apply a smoothing operation to this tensor by convolving it with a predefined Gaussian kernel denoted as $K$. In practice, we have to construct an inverse pixel repository as
\begin{equation}
\label{eq:inverse-coding}
\mathcal{\tilde{R}}_c = \{(i, y, x) \mid \mathbf{X}_{\mathcal{E}}[i, y, x] = c\}
\end{equation}
that stores the coordinates of pixels associated with each code $c$ where $i$ indicates the frame index, and $y$ and $x$ represent the 2D spatial coordinates. With the inverse repository, we can formally define the local harmonization as\footnote{$\mathcal{R}^t_c$ no longer represents a single shared pixel but a tensor of length $|\mathcal{\tilde{R}}_c|$ here. We reuse the notation for convenience.}
\begin{equation}
\label{eq:local-harmonization-solver}
\mathcal{R}^t_c = G_{conv}\left(\mathbf{X}^t\right) = \mathcal{P}\left(\mathbf{X}^t\left[\mathcal{\tilde{R}}^t_c\right]\right) * K, \; \forall c \in [0, n),
\end{equation}
where $\mathcal{P}$ is a reflection padding function to maintain the size, $|\mathbf{X}^t[\mathcal{\tilde{R}}^t_c]|$, after convolution and permit batch implementations. A drawback to this approach is that Equation~\ref{eq:inverse-coding} cannot be executed in parallel and has a time complexity of $\mathcal{O}(\mathcal{T}HW)$, i.e. the number of pixels in $\mathbf{X}$. The local harmonization approach is not physically correct but it provides a fluent transition of color in time that is visually appealing.

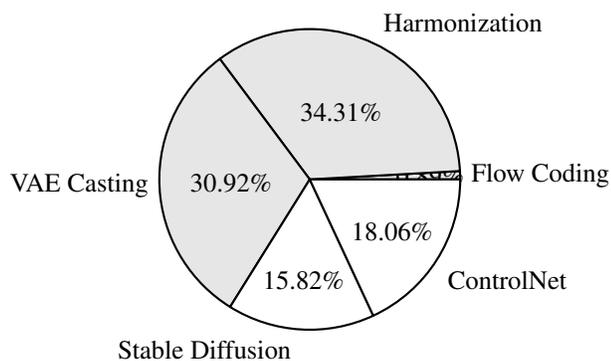
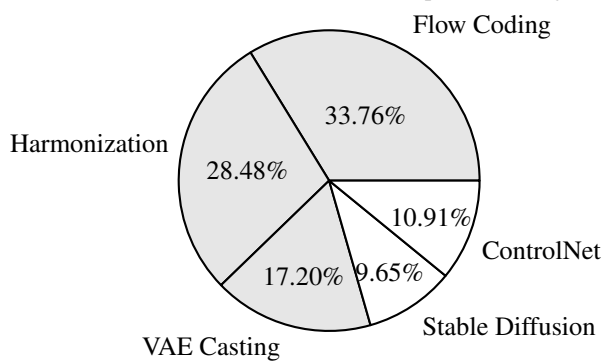
\begin{figure}[t]
\centering
\begin{subfigure}{\linewidth}
\begin{tikzpicture}
\pie[radius=2,color={black!10,black!10,black!10,white,white}]{
0.89/Flow Coding,
34.31/Harmonization,
30.92/VAE Casting,
15.82/Stable Diffusion,
18.06/ControlNet
}
\end{tikzpicture}
\caption{Global harmonization. \textbf{6.8} seconds per frame (avg.)}
\label{fig:pie-gh}
\end{subfigure}
\begin{subfigure}{\linewidth}
\begin{tikzpicture}
\pie[radius=2,color={black!10,black!10,black!10,white,white}]{
33.76/Flow Coding,
28.48/Harmonization,
17.20/VAE Casting,
9.65/Stable Diffusion,
10.91/ControlNet
}
\end{tikzpicture}
\caption{Local harmonization. \textbf{11.9} seconds per frame (avg.)}
\label{fig:pie-lh}
\end{subfigure}
\caption{\textbf{Execution time distribution for generating videos in the entire MPI Sintel.} The components proposed in this work are shaded.}
\label{fig:exe-time}
\end{figure}

\paragraph{Execution time.} We provide an analysis of execution time required for both global and local harmonization in Figure \ref{fig:exe-time}. The results are collected by generating the entire MPI Sintel dataset with GT optical flows. The insights from Figure \ref{fig:pie-gh} demonstrate that the total time taken for video synthesis using global harmonization is merely three times that of employing the original Stable Diffusion in combination with ControlNet. Moreover, Figure \ref{fig:pie-lh} underscores that in the case of local harmonization, both the time taken for initial coding (building reverse repository) and the subsequent harmonization (convolution) in each diffusion step are notably extended. As a result, the execution time for the local harmonization approach is almost five times that of utilizing the original Stable Diffusion combined with ControlNet.

\subsection{Experiments}
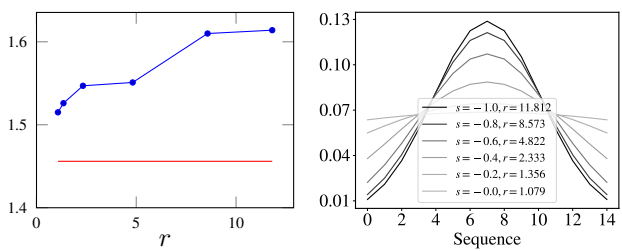
\begin{figure}[t]
\centering
\begin{subfigure}{0.49\linewidth}
\centering
\begin{tikzpicture}
\pgfplotsset{every tick label/.append style={font=\tiny}}
\pgfplotstableread[col sep=comma]{conv-epe.csv}\datatable
\begin{axis}[
ymin=1.4,
xlabel={$r$},
xticklabel={\pgfmathprintnumber[assume math mode=true]{\tick}},
yticklabel={\pgfmathprintnumber[assume math mode=true]{\tick}},
x label style={at={(axis description cs:0.5,-0.09)},anchor=north},
width=5cm, 
height=4.18cm 
]
\addplot+[mark size=1pt] table[x={r}, y={epe}] {\datatable};
\addplot+[mark=none] table[x={r}, y={epe-gh}] {\datatable};
\end{axis}
\end{tikzpicture}
\caption{EPE for local harmonization}
\label{fig:conv-epe}
\end{subfigure}
\hfill
\begin{subfigure}{0.49\linewidth}
\centering
\includegraphics[width=\linewidth]{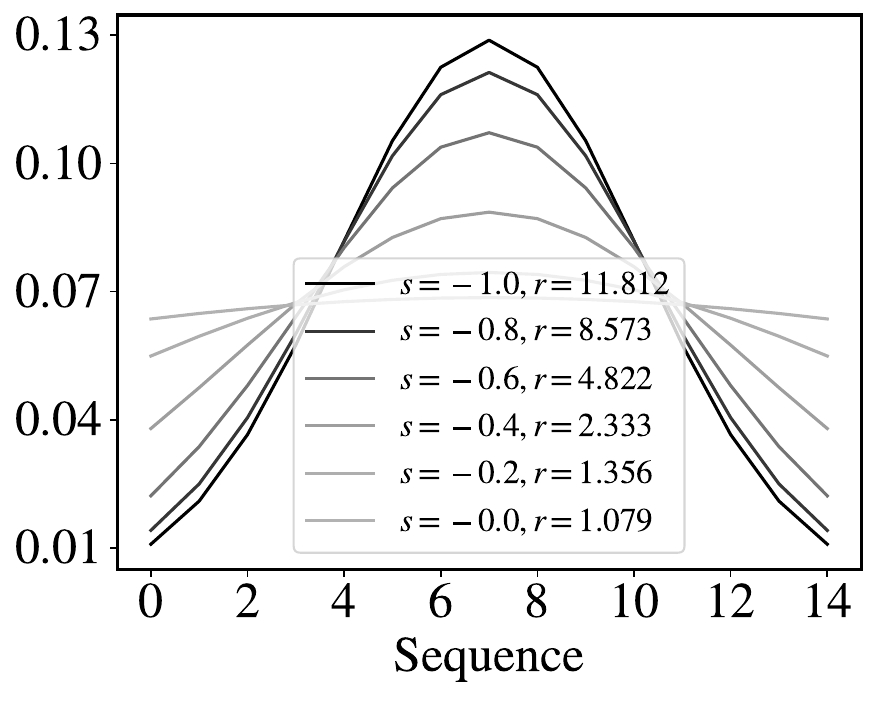}
\caption{Convolution kernels}
\label{fig:stds}
\end{subfigure}
\caption{\textbf{Local harmonization results on MPI Sintel.} The red line in \ref{fig:conv-epe} is the EPE of global harmonization.}
\label{fig:local-harmonization-exp}
\end{figure}

In this section, we conduct a quantitative assessment of the efficacy of local harmonization using various Gaussian kernels. The discrete 1D Gaussian kernel $K$, characterized by a width $\ell$, exhibits an impulse response defined by
\begin{equation}
f(x) = \frac{1}{\sigma\sqrt{2\pi}}\exp\left( -\frac{1}{2}\left(\frac{x-\ell / 2}{\sigma}\right)^{\!2}\,\right), \; x \in \mathbb{Z} \cap [0, \ell],
\end{equation}
where $\sigma$ denotes the standard deviation regulating the kernel's sharpness. Prior to application, we normalize the kernel to ensure its weights sum to one.  We control the harmonization strength by varying $\sigma$. To obtain a proper scale of $\sigma$, we define a set of seeds $s$ ranging from $-1.0$ to $0.0$ at intervals of $0.2$, and set $\sigma = 100^s + 0.2$. Additionally, $r$ signifies the ratio of the maximum weight to the minimum weight in $K$, serving as an intuitive indicator of the convolutional harmonization's smoothing level, as illustrated in Figure \ref{fig:stds}. We graph the End Point Error (EPE) of generated videos against $r$ in Figure \ref{fig:conv-epe}. The outcomes affirm that videos exhibit enhanced fluidity as $r$ diminishes, albeit at the cost of a more rigorous global consistency constraint.

\section{User Study}
\begin{figure}
    \centering
    \includegraphics[width=0.95\linewidth]{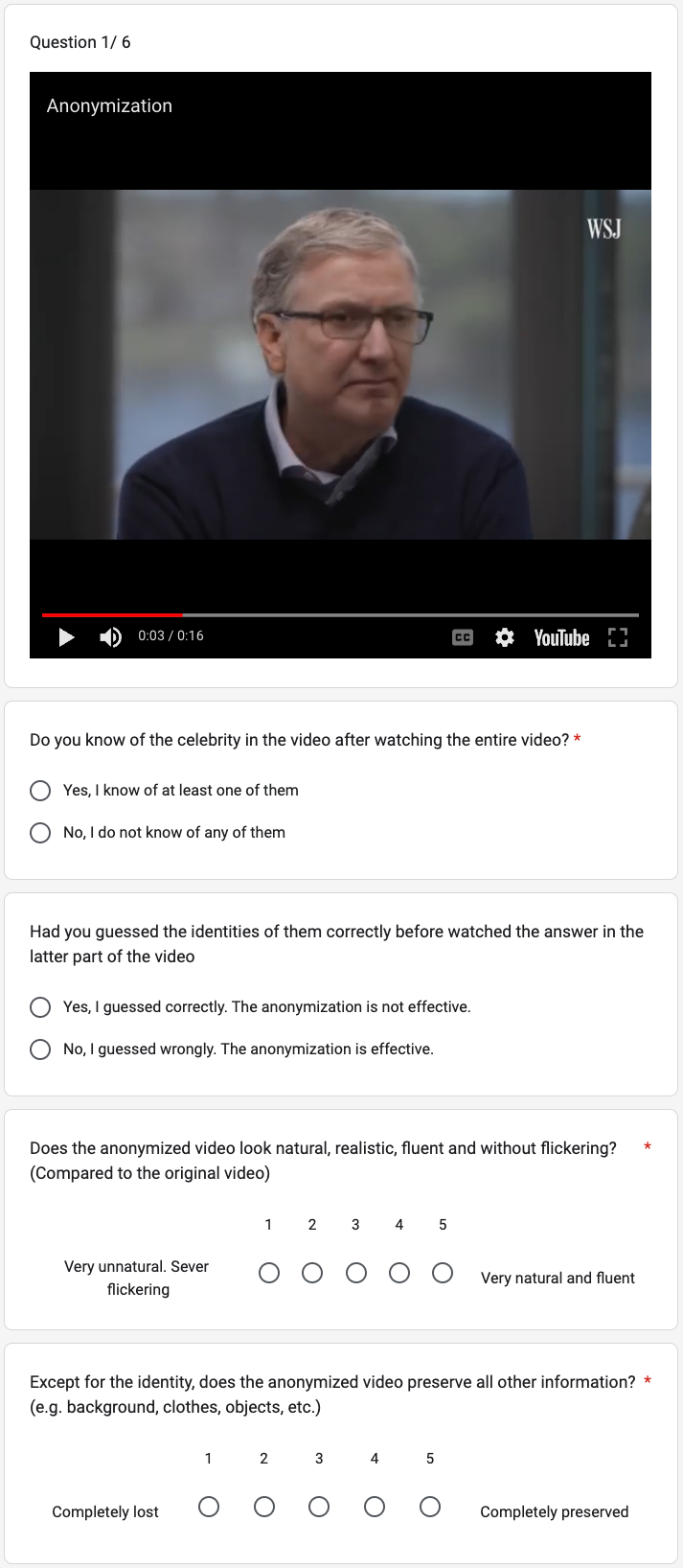}
    \caption{\textbf{Screenshot of the survey for anonymization comparison.} In the video, we first show the anonymized version to the participants, then we show the original version. We have made sure that the participants know of the celebrity before taking their responses into account. More videos can be found in the supplementary materials.}
    \label{fig:survey}
\end{figure}

We include a representative screenshot as an illustration of our user study in Figure \ref{fig:survey}. Our survey on anonymization is conducted interactively and possesses a unique aspect: each participant encounters it only once. This survey pertains to guessing the identity concealed within the anonymized rendition of each video. Notably, participants are provided no foreknowledge or contextual information about the celebrity under consideration. This characteristic mandates that the source videos utilized for assessing various methods cannot be reused.

\section{Additional Samples}
\begin{figure*}
\centering
\begin{tabular}{cccc}
\includegraphics[width=0.24\linewidth]{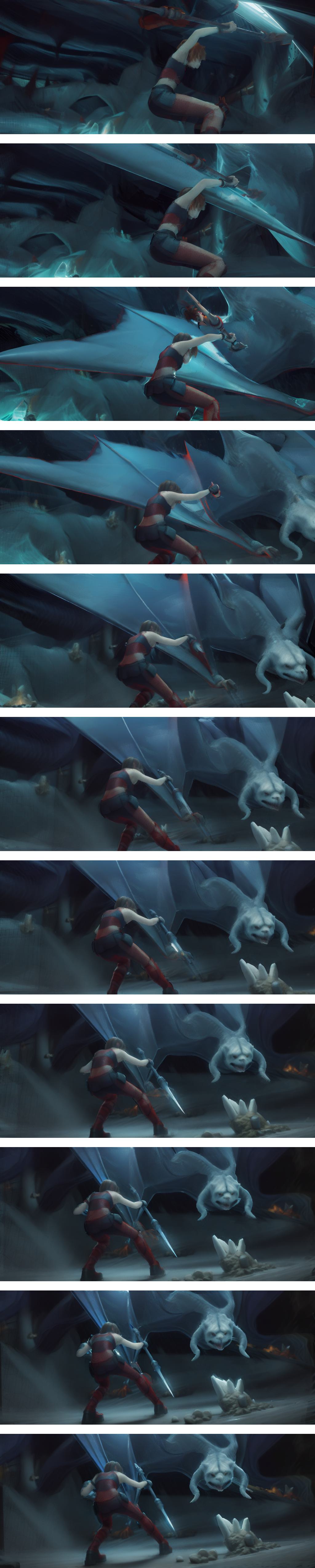} &
\includegraphics[width=0.24\linewidth]{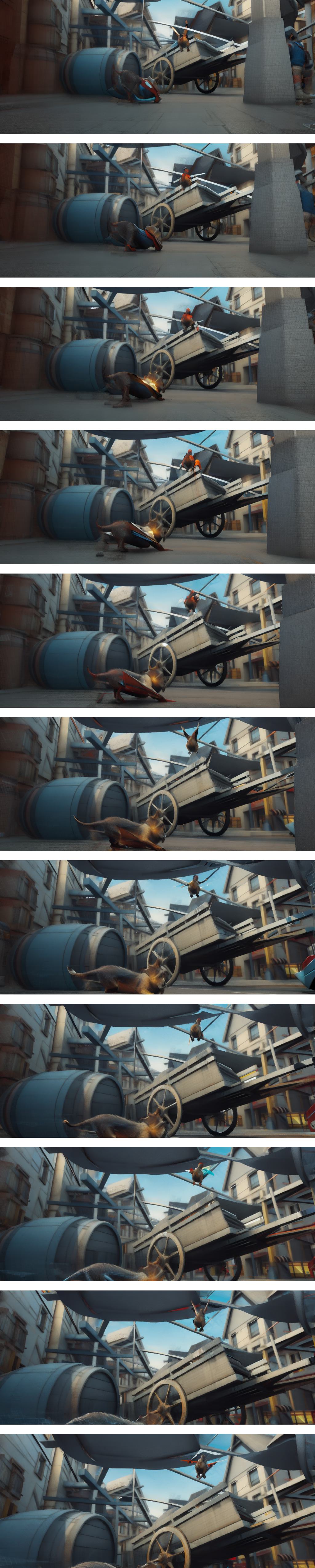} &
\includegraphics[width=0.24\linewidth]{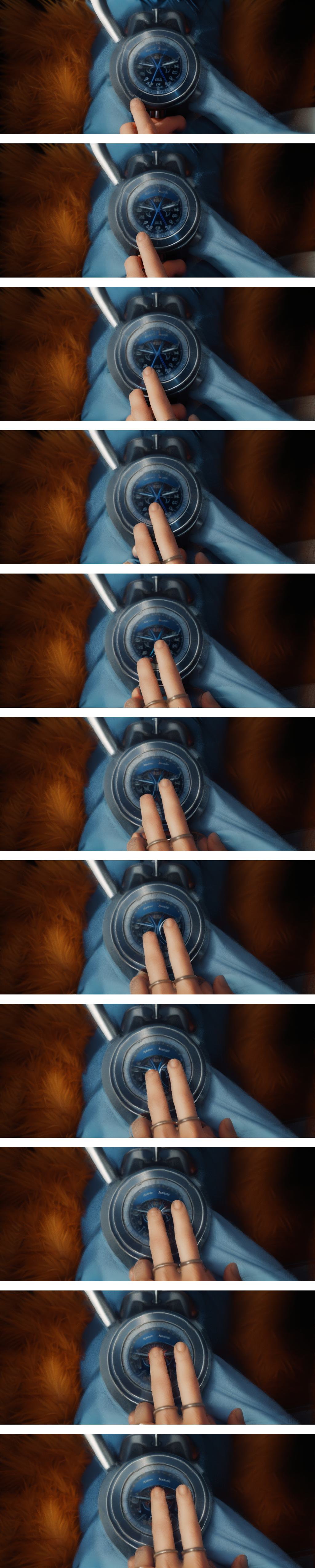} &
\includegraphics[width=0.24\linewidth]{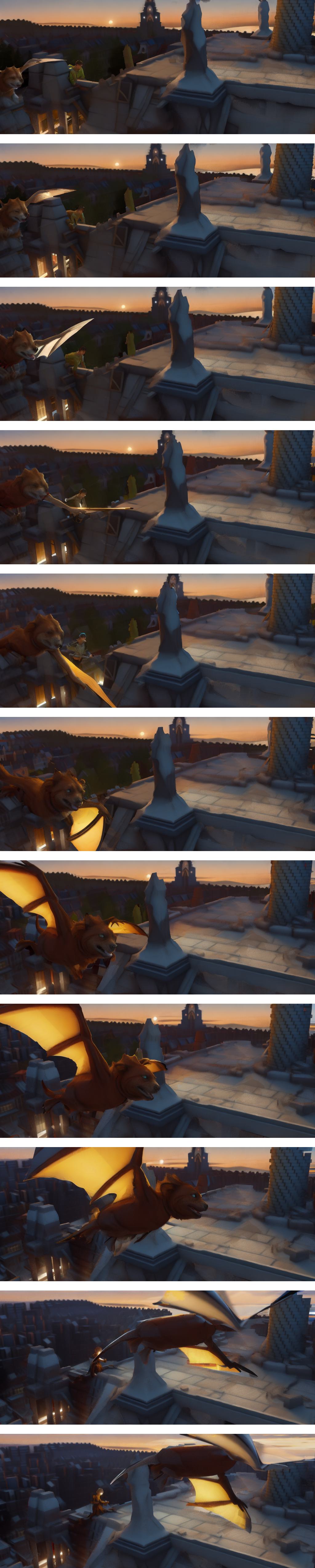}
\end{tabular}
\caption{\textbf{Samples of video rendering on MPI Sintel.} The videos are generated using the lineart ControlNet from pure noise.}
\label{fig:appendix-render}
\end{figure*}

\begin{figure*}
\centering
\begin{tabular}{cccc}
\includegraphics[width=0.24\linewidth]{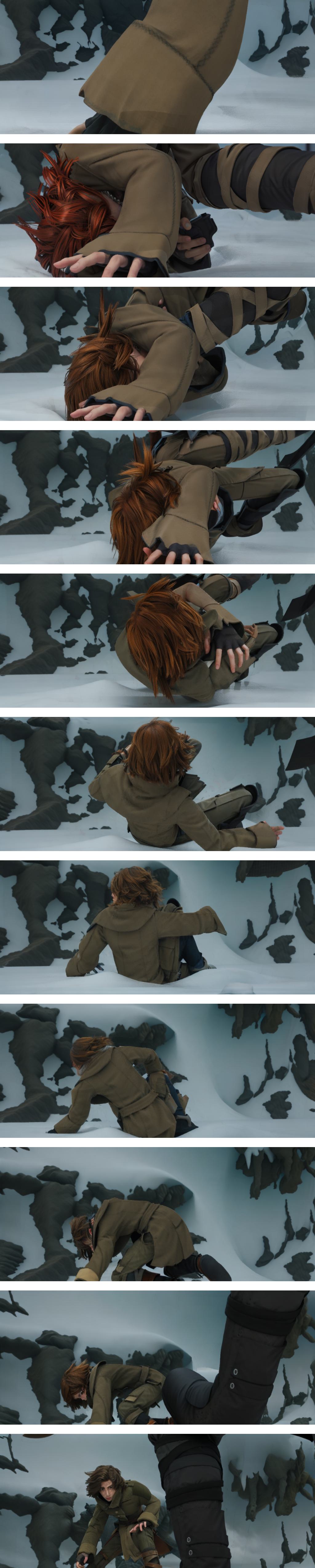} &
\includegraphics[width=0.24\linewidth]{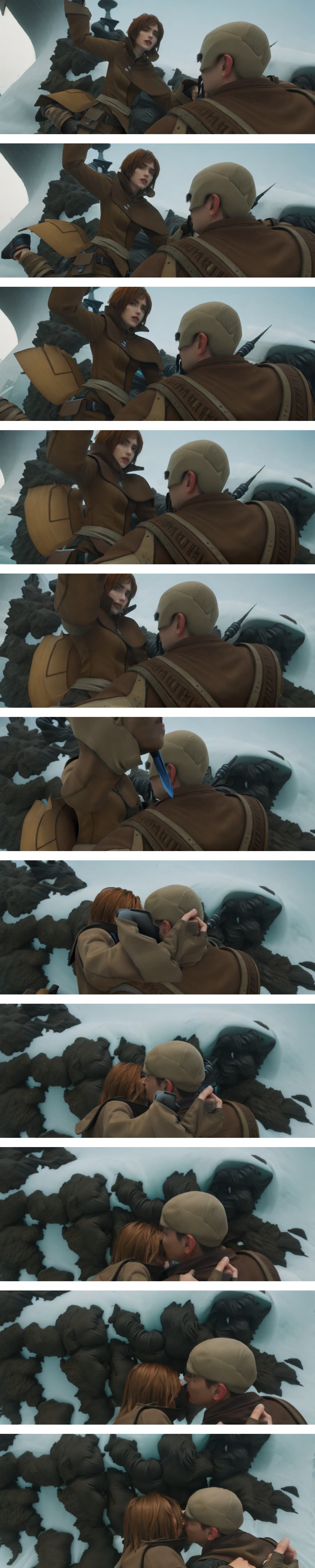} &
\includegraphics[width=0.24\linewidth]{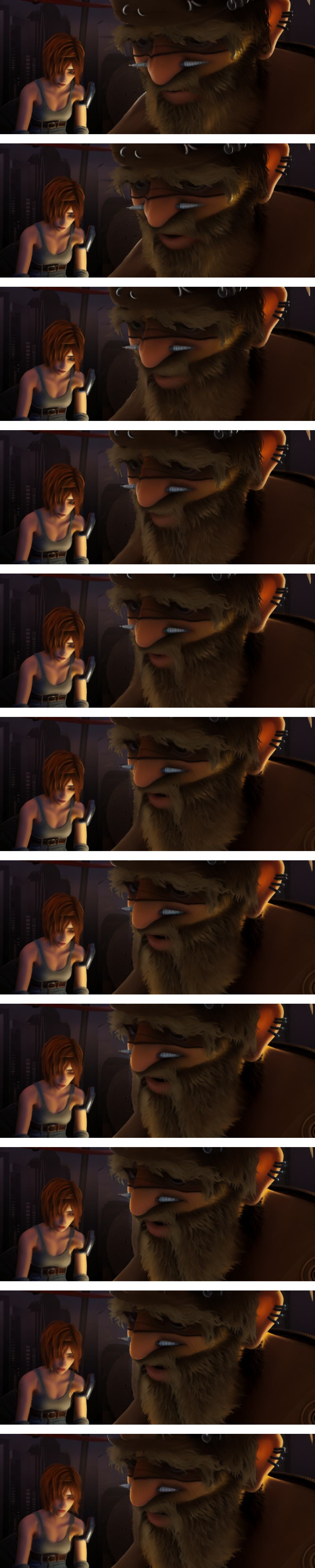} &
\includegraphics[width=0.24\linewidth]{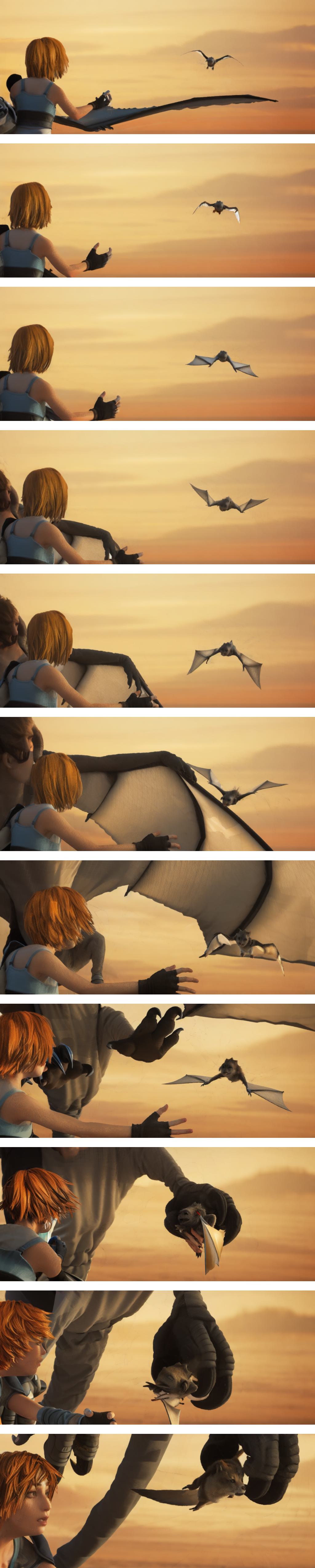}
\end{tabular}
\caption{\textbf{Samples of assistive video rendering on MPI Sintel.} The videos are generated using the lineart ControlNet with a noise level of $0.5T$.}
\label{fig:appendix-assistive-render}
\end{figure*}

We provide additional visual samples of our generated videos in Figure \ref{fig:appendix-render} and Figure \ref{fig:appendix-assistive-render}.

\clearpage

\setlength\tabcolsep{2pt}

\end{document}